\documentclass{article}

\usepackage{microtype}
\usepackage{graphicx}
\usepackage{subfigure}
\usepackage{booktabs} 
\usepackage[inline]{enumitem}
\usepackage{multirow}
\usepackage{algorithm}
\usepackage{algorithmic}
\usepackage[algo2e, vlined, ruled, linesnumbered]{algorithm2e}

\usepackage{graphbox}

\usepackage{mathtools}

\usepackage{Definitions}
\usepackage{hyperref}
\hypersetup{colorlinks, citecolor={blue}}

\usepackage[accepted]{icml2020}

\newcommand{\modelshort}{Retro*}

\usepackage{xr-hyper}
\makeatletter
\newcommand*{\addFileDependency}[1]{
  \typeout{(#1)}
  \@addtofilelist{#1}
  \IfFileExists{#1}{}{\typeout{No file #1.}}
}
\makeatother
 
\newcommand*{\myexternaldocument}[1]{%
    \externaldocument{#1}%
    \addFileDependency{#1.tex}%
    \addFileDependency{#1.aux}%
}

\myexternaldocument{appendix}

\icmltitlerunning{\modelshort{}: Learning Retrosynthetic Planning with Neural Guided A* Search}

\begin{document}

\twocolumn[
\icmltitle{\modelshort{}: Learning Retrosynthetic Planning with Neural Guided A* Search}



\icmlsetsymbol{equal}{*}

\begin{icmlauthorlist}
\icmlauthor{Binghong Chen}{gatech}
\icmlauthor{Chengtao Li}{galixir}
\icmlauthor{Hanjun Dai}{google}
\icmlauthor{Le Song}{gatech,ant}
\end{icmlauthorlist}

\icmlaffiliation{gatech}{College of Computing, Georgia Institute of Technology}
\icmlaffiliation{galixir}{Galixir}
\icmlaffiliation{google}{Google Research, Brain Team}
\icmlaffiliation{ant}{Ant Financial}

\icmlcorrespondingauthor{Binghong Chen}{binghong@gatech.edu}

\icmlkeywords{Machine Learning, ICML}

\vskip 0.3in
]



\printAffiliationsAndNotice{}  

\begin{abstract}
Retrosynthetic planning is a critical task in organic chemistry which identifies a series of reactions that can lead to the synthesis of a target product. The vast number of possible chemical transformations makes the size of the search space very big, and retrosynthetic planning is challenging even for experienced chemists. However, existing methods either require expensive return estimation by rollout with high variance, or optimize for search speed rather than the quality.  In this paper, we propose Retro*, a neural-based A*-like algorithm that finds high-quality synthetic routes efficiently. It maintains the search as an AND-OR tree, and learns a neural search bias with off-policy data. Then guided by this neural network, it performs best-first search efficiently during new planning episodes. Experiments on benchmark USPTO datasets show that, our proposed method outperforms existing state-of-the-art with respect to both the success rate and solution quality, while being more efficient at the same time.
\end{abstract}

\setlength{\abovedisplayskip}{2pt}
\setlength{\abovedisplayshortskip}{2pt}
\setlength{\belowdisplayskip}{2pt}
\setlength{\belowdisplayshortskip}{2pt}
\setlength{\jot}{2pt}

\setlength{\floatsep}{2ex}
\setlength{\textfloatsep}{2ex}

\section{Introduction}

Retrosynthetic planning is one of the fundamental problems in organic chemistry. Given a target product, the goal of retrosynthesis is to identify a series of reactions that can lead to the synthesis of the product, by searching backwards and iteratively applying chemical transformations to unavailable molecules. As thousands of theoretically-possible transformations can all be applied during each step of reactions, the search space of planning will be huge and makes the problem challenging even for experienced chemists.

The one-step retrosynthesis prediction, which predicts a list of possible direct reactants given product, serves as the foundation for realizing the multistep retrosynthetic planning. Existing methods roughly fall into two categories, either template-based or template-free. Each chemical reaction is associated with a reaction template that encodes how atoms and bonds change during the reaction. Given a target product, template-based methods predict the possible reaction templates, and subsequently apply the predicted reaction templates to target molecule to get corresponding reactants. Existing methods include retrosim~\citep{coley2017computer}, neuralsym~\cite{segler2017neural} and GLN~\citep{dai2019retrosynthesis}.  Though conceptually straightforward, template-based methods need to deal with tens or even hundreds of thousands of possible reaction templates, making the classification task hard. Besides, templates are not always available for chemical reactions. Due to these reasons, people have also been developing template-free methods that could directly predict reactants. Most of existing methods employ seq2seq models like LSTM~\citep{liu2017retrosynthetic} or Transformer~\citep{karpov2019transformer} from neural machine translation literature. 

While one-step methods are continuously being improved, most molecules in real world cannot be synthesized within one step. Possible number of synthesis steps could go up to 60 or even more. Since each molecule could be synthesized by hundreds of different possible reactants, the possible synthesis routes becomes countless for a single product. Such huge space poses challenges for efficient searching and planning, even with advanced one-step approaches. 

Besides the huge search space, another challenge is the ambiguity in performance measure and benchmarking. It has been extremely hard to quantitatively analyze the performance of any multi-step retrosynthesis algorithms due to the ambiguous definition of `good synthesis routes', nor are there any benchmark datasets for analyzing designed algorithms. 
Most common ways for quantitative analysis is to employ domain experts and let them judge if one synthesis route is better than the other based solely on their experiences, which is both time-consuming and costly.

Due to aforementioned challenges, there are less work proposed in the field of multi-step retrosynthetic planning. Previous works using Monte Carlo Tree Search (MCTS)~\citep{segler2018planning, segler2017towards} have achieved superior results over neural- or heuristic-based Breadth First Search (BFS). However, MCTS-based methods has several limitations in this setting:  
\begin{itemize}[leftmargin=*,nolistsep,nosep]
    \item Each tree node corresponds to a set of molecules instead of single molecule. This addtional combinatorial aspect make the representation of tree node, and the estimation of its value even harder. Furthermore, reactions do not explicilty appear as nodes in the tree, which prevents their algorithm from exploiting the structure of subproblems. 
    \item As the algorithm depends on online value estimation, the full rollout from vanilla MCTS may not be efficient for the planning need. Furthermore, the algorithm can not exploit historical data in that many good retrosysthesis plans may have been found previously, and ``intuitions'' on how to plan efficiently may be learned from these histories.  
\end{itemize}
For quantitative evaluation, they have employed numerous domain experts to conduct A-B tests over methods proposed by their algorithm and other baselines. 

In this paper, we present a novel neural-guided tree search method, called Retro*\footnote{Available at \url{https://github.com/binghong-ml/retro_star}}, for chemical retrosynthesis planning. In our method, 
\begin{itemize}[leftmargin=*,nolistsep,nosep]
    \item We explicitly maintain information about reactions as nodes in an AND-OR tree, where a node with ``AND'' type corresponds to a reaction, and a node with ``OR'' type corresponds to a molecule. The tree captures the relations between candidate reactions and reactant molecules, which allows us to exploit structure of subproblems corresponding to a single molecule. 
    \item Based on the AND-OR tree representation, we propose an A*-like planning algorithm which is guided by a neural network learned from past retrosynthesis planning experiences. More specifially, The neural network learns a synthesis cost for each molecule, and it helps the search algorithm to pick the most promising molecule node to expand.
\end{itemize}
Furthermore, we also propose a method for constructing benchmark synthesis routes data given reactions and chemical building blocks. Based on this, we construct a synthesis route dataset from benchmark reaction dataset USPTO. The route dataset is not only useful for quantitative analysis for predicted synthesis routes, but also work as training data for the neural network components in our method. 

Below we summarize our contributions:
\begin{itemize}[leftmargin=*,nolistsep,nosep]
    \item We propose a novel learning-based retrosynthetic planning algorithm to learn from previous planning experience.
    The proposed algorithm outperforms state-of-the-art methods by a large margin on a realworld benchmark dataset.
    \item Our algorithm framework can induce a search algorithm that guarantees the optimal solution. 
    \item We propose a method for constructing synthesis route datasets for quantitative analysis of multistep retorsynthetic planning methods.
\end{itemize}

Our planning algorithm is general in the sense that it can also be applied to other machine learning problems such as theorem proving~\citep{yang2019learning} and hierarchical task planning~\citep{erol1996hierarchical}. A synthetic task planning experiment is included in Appendix~\ref{sec:htn} to demonstrate the idea. Most related works have been mentioned in the first two sections. For more related works, please refer to Appendix~\ref{sec:related}.





\section{Background}

In this section, we first state the problem and its background we are tackling in~\secref{sec:prob_stmt}. Then in~\secref{sec:mcts} and~\secref{sec:pns} we describe how MCTS and proof number search fit in the problem setting.

\subsection{Problem Statement}
\label{sec:prob_stmt}

\noindent\textbf{One-step retrosynthesis:} Denote the space of all molecule as $\Mcal$. The one-step retrosynthesis takes a target molecule $t \in \Mcal$ as input, and predicts a set of source reactants $\Scal \subset \Mcal$ that can be used to synthesize $t$. This is the reverse problem of reaction outcome prediction. In our paper, we assume the existence of such one-step retrosynthesis model (or one-step model for simplicity in the rest of the paper) $B$, 
\begin{equation}
\textstyle
	B(\cdot):\quad t \rightarrow \{R_i, \Scal_i, c(R_i)\}_{i=1}^k 
\end{equation}
which outputs at most $k$ reactions $R_i$, the corresponding reactant sets $\Scal_i$ and costs $c(R_i)$. The cost can be the actual price of the reaction $R_i$, or simply the negative log-likelihood of this reaction under model $B$. A one-step retrosynthesis model can be learned from a dataset of chemical reactions $\Dcal_{train} = \cbr{\Scal_i, t_i}$~\footnote{For simplicity we follow the common practice to ignore the reagents and other chemical reaction conditions.} which have already been discovered by chemists in the past~\cite{coley2017computer, segler2017neural, liu2017retrosynthetic, dai2019retrosynthesis, karpov2019transformer}. 

\noindent\textbf{Retrosynthesis planning.}
Given a single target molecule $t \in \Mcal$ and an initial set of molecules $\Ical \subset \Mcal$, we are interested in synthesizing $t$ via a sequence of chemical reactions using reactants that are from or can be synthesized by $\Ical$. In this case, $\Ical$ corresponds to a set of molecules that are commercially available. The goal of retrosynthesis planning is to predict a sequence of reactions with reactants in $\Ical$ and will ultimately arrive at product $t$. 

Instead of performing forward chaining like reasoning that starts from $\Ical$, a more efficient and commonly used method is to perform backward chaining that starts from the molecule $t$, and perform a series of one-step retrosynthesis prediction until all the reactants required are from $\Ical$. Beyond just finding such a synthesis route, our goal is to find the retrosynthesis plan that are:
\begin{itemize}[leftmargin=*,nolistsep,nosep]
\item High-quality: 
    \begin{itemize}
        \item The entire retrosynthesis plan should be chemically sound with high probability;
        \item The reactants or chemical reactions required should have as low cost as possible;
    \end{itemize}
\item Efficient: Due to the synthesis effort, the number of retrosynthesis steps should be limited.
\end{itemize}
Our proposed \modelshort{} is aiming at finding the best retrosynthesis plan with respect to above criteria in limited time. To achieve this, we also assume that the quality of a solution can be measured by the reaction cost, where such cost is known to our model. 

\subsection{Monte Carlo Tree Search}
\label{sec:mcts}


\begin{figure}[ht]
\centering
\includegraphics[width=0.43\textwidth]{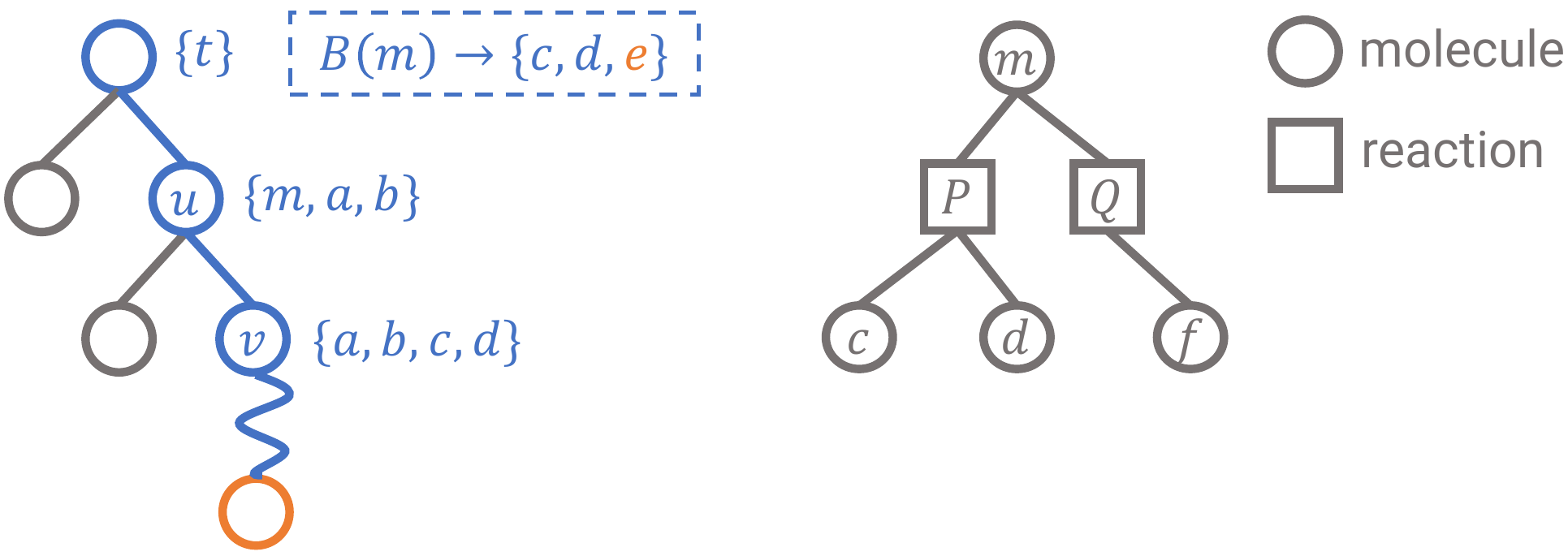}
\caption{\textbf{Left}: MCTS~\citep{segler2018planning} for retrosynthesis planning. Each node represents a set of molecules. Orange nodes/molecules are available building blocks; \textbf{Right}: AND-OR stump illustration of $B(m) = {P, Q}$. Reaction $P$ requires molecule $c$ and $d$. Reaction $Q$ requires molecule $f$. Either $P$ or $Q$ can be used to synthesize $m$. }
\label{fig:mcts-and-or-stump}
\end{figure}

The Monte Carlo Tree Search (MCTS) has achieved ground breaking successes in two player games, such as GO~\citep{silver2016mastering, silver2017mastering}. Its variant, UCT~\citep{kocsis2006bandit}, is especially powerful for balancing exploration and exploitation in online learning setting, and has been employed in~\citet{segler2018planning} for retrosynthesis planning. Specifically, as illustrated in~\figref{fig:mcts-and-or-stump}, the tree search start from the target molecule $t$. Each node $u$ in the current search tree $T$ represents a set of molecules $\Mcal_u$. Each child node $v \in ch(u)$ of $u$ is obtained by selecting one molecule $m \in \Mcal_u$ and a one-step retrosynthesis reaction $\rbr{R_{uv}, \Scal_{uv}, c\rbr{R_{uv}}} \in B(m)$, where the resulting node $v$ contains molecule set $\Mcal_v = (\Scal_{uv} \cup \Mcal_u) \setminus \cbr{m} \setminus \Ical$. 

Despite its good performance, MCTS formulation for retrosynthesis planning has several limitations. First, the rollout needed in MCTS makes it time-consuming, and unlike in two-player zero-sum games, the retrosynthesis planning is essentially a single player game where the return estimated by random rollouts could be highly inaccurate. Second, since each tree node is a set of molecules instead of a single molecule, the combinatorial nature of this representation brings the sparsity in the variance estimation.

\subsection{Proof Number Search and Variants}
\label{sec:pns}


The proof-number search (PNS)~\citep{allis1994proof} is a game tree search that is designed for two-player game with binary goal. It tries to either prove or disprove the root node as fast as possible. In the retrosynthesis planning scenario, this corresponds to either proving the target molecule $t$ by finding a feasible planning path, or concluding that it is not synthesizable.  

\noindent\textbf{AND-OR Tree:} The search tree of PNS is an AND-OR tree $T$, where each AND node needs all its children to be proved, while OR node requires at least one to be satisfied. Each node $u \in T$ is associated with a proof number $pn(u)$ that defines the minimum number of leaf nodes to be proved in order to prove $u$. Similarly, the disproof number $dn(u)$ finds the minimum number of leaf nodes needed to disprove $u$. With such definition, we can recursively define these numbers for internal nodes. Specifically, for AND node $u$, 
\begin{eqnarray}
\textstyle
& pn(u) = \sum_{v \in ch(u) }pn(v), dn(u) = \min_{v \in ch(u)} dn(v) \nonumber \\
& \text{and for proved nodes: } pn(u) = 0, dn(u) = +\infty
\end{eqnarray}
and for OR node $u$, we have
\begin{eqnarray}
\textstyle
& pn(u) = \min_{v \in ch(u) }pn(v), dn(u) = \sum_{v \in ch(u)} dn(v) \nonumber \\
& \text{and for disproved node: } pn(u) = +\infty, dn(u) = 0 
\label{eq:pn_or}
\end{eqnarray}

{\bf Represent retrosynthesis planning using AND-OR tree:} As illustrated in~\figref{fig:mcts-and-or-stump}, the application of one-step retrosynthesis model $B$ on molecule $m$ can be represented using one block of AND-OR tree (denoted as AND-OR stump), with molecule node as `OR' node and reaction node as `AND' node. This is because a molecule $m$ can be synthesized using \underline{\textbf{any one}} of its children reactions (or-relation), and each reaction node requires \underline{\textbf{all}} of its children molecules (and-relation) to be ready.  

The search of PNS starts from the root node every time, and selects the child node with either minimum proof number or minimum disproof number, depends on whether the current node is an OR node or AND node, respectively. The process ends when a leaf node is reached, which can be either reaction or molecule node to be expanded. And after one step of retrosynthesis expansion, all the $pn(\cdot)$ and $dn(\cdot)$ of nodes along the path back to the root will be updated. The two-player game in this sense comes from the interleaving behavior of selecting proof and disproof numbers, where the first `player' tries to prove the root while the second `player' tries to disprove it. As both of the players behave optimally when the proof/disproof numbers are accurate, such perspective would bring the efficiency for finding a feasible synthesis path or prove that it is not synthesizable.

{\bf Variant:} There have been several variants to improve different aspects of PNS, including different traversal strategy, different initialization methods of $pn(\cdot)$ and $dn(\cdot)$ for newly added nodes. The most recent work DFPN-E~\citep{kishimoto2019depth} builds on top of the depth-first variant of PNS with an additive cost in addition to classical update rule in Eq~\eqref{eq:pn_or}. Specifically, for an unsolved OR node, 
\begin{equation}
	pn(u) = \min_{v \in ch(u)} \rbr{h(u, v) + pn(v)}
\end{equation} 
Here $h(u, v)$ is the function of the cost of corresponding one-step retrosynthesis. Together with manually defined thresholds, this method addresses the \textit{lopsided} problem in retrosynthesis planning, \ie, the imbalance of branching factor between AND and OR nodes.

The variants of PNS has shown some promising results over MCTS for retrosynthesis planning. However, the two-player game formulation is designed for the speed of a proof, not necessarily the overall solution quality. Moreover, existing works rely on human expert to design $pn(\cdot), dn(\cdot)$ and thresholds during search. This makes it not only time-consuming to tune, but also hard to generalize well when solving new target molecule $t$ or dealning with new one-step model or reaction data.

\section{\modelshort{} Search Algorithm}

\begin{figure*}[ht]
\vspace{-1mm}
\centering
\includegraphics[width=0.8\textwidth]{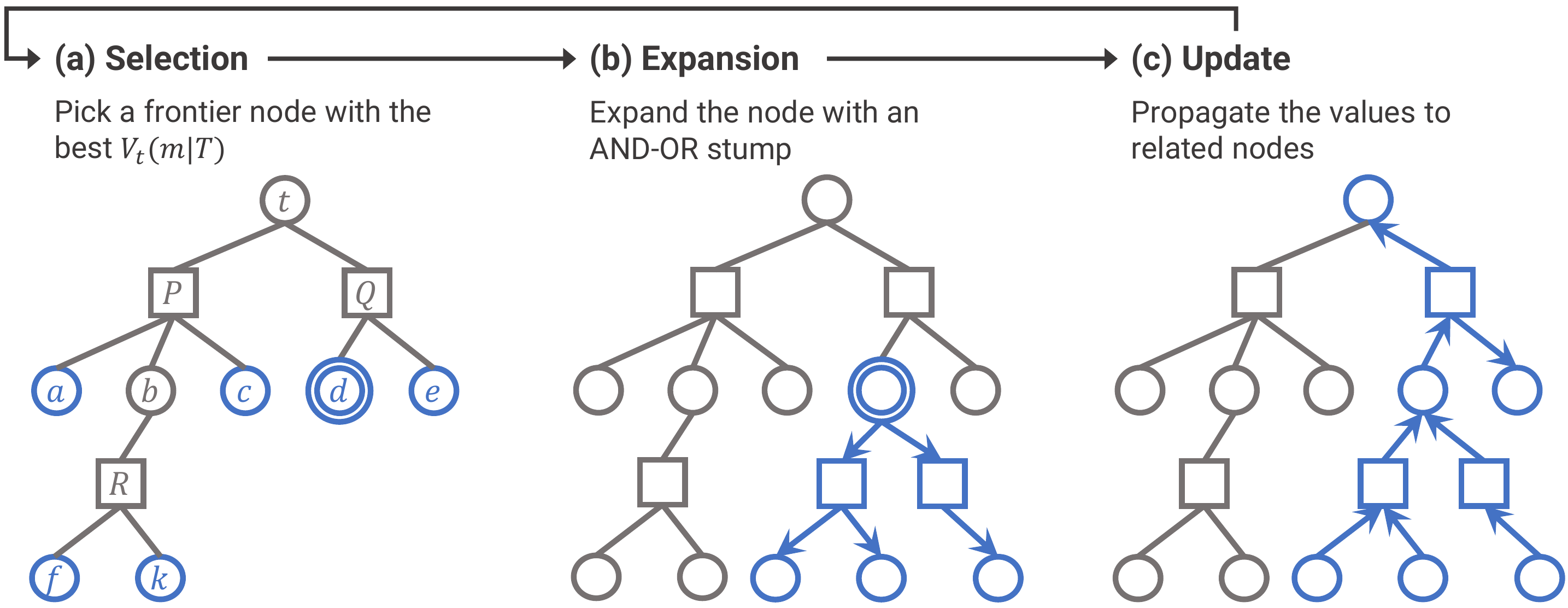}
\vspace{-2mm}
\caption{\modelshort{} algorithm framework. We use circles to represent molecule nodes, and squares to represent reaction nodes. An iteration consists of three phases. In the selection phase, one of the frontier molecule nodes is selected according to the cost estimation $V_t(m|T)$. Then the an AND-OR stump is expanded from the selected node. All the new reactions and molecules are added to the tree. Finally the values inside the tree are updated using the $V_m$s from the newly added molecules.
The left-most figure also serves as the illustration for computing $V_t(f|T)$. $V_t(f|T) = g_t(f|T) + h_t(f|T)$, where $g_t(f|T) = c(P) + c(R)$, and $h_t(f|T) = V_a + V_c + V_f + V_k$.} 
\vspace{-2mm}
\label{fig:alg-framework}
\end{figure*}

\begin{algorithm2e}[t]
Initialize $T = (\Vcal, \Ecal)$ with $\Vcal\gets \{t\}$, $\Ecal\gets\emptyset$\;
\While{route not found}{
    $m_{next} \gets \argmin_{m\in\Fcal(T)} V_t(m)$\;
    $\{R_i, \Scal_i, c(R_i)\}_{i=1}^k \gets B(m_{next})$\;
    \For{$i\gets1$ \KwTo $k$}{
        Add $R_i$ to $T$ under $m_{next}$\;
        \For{$j\gets1$ \KwTo $|\Scal_i|$}{
            Add $\Scal_{ij}$ to $T$ under $R_i$\;
            
        }
    }
    Update $V_t(m)$ for $m$ in $\Fcal(T)$\;
}

\KwRet route\;
\caption{$\mathtt{Retro^*}(t)$}
\label{alg:main}
\end{algorithm2e}

Our proposed \modelshort{} is a retrosynthetic planning algorithm that works on the AND-OR search tree. It is significantly different from PNS which is also based on AND-OR tree, or other MCTS based methods in the following ways: 
\begin{itemize}[leftmargin=*,nolistsep,nosep]
\item \modelshort{} utilizes AND-OR tree for \textit{single} player game which only utilizes the global value estimation. This is different from PNS which models the problem as \textit{two-player} game with both proof numbers and disproof numbers. The distinction of the objective makes \modelshort{} advantageous in finding best retrosynthetic routes.

\item \modelshort{} estimates the future value of frontier nodes with neural network that can be trained using historical retrosynthesis planning data. This is different from the expensive rollouts used in~\citet{segler2018planning}, or the human designed heuristics in~\citet{kishimoto2019depth}. This not only enables more accurate prediction during expansion, but also generalizes the knowledge learned from existing planning paths.
\end{itemize}

\subsection{Overview of \modelshort{}}

\modelshort{}~(\algref{alg:main}) is a best-first search algorithm, which exploits neural priors to directly optimize for the quality of the solution. The search tree $T$ is an AND-OR tree, with molecule node as 'OR' node and reaction node as 'AND' node. 
It starts the search tree $T$ with a single root molecule node that is the target molecule $t$. At each step, it selects a node $u$ in the frontier of $T$ (denoted as $\Fcal(T)$) according to the value function. Then it expands $u$ with the one-step model $B(u)$ and grows $T$ with one AND-OR stump. Finally the nodes with potential dependency on $u$ will be updated. Below we first provide a big picture of the algorithm by explaining these steps one by one, then we look into details of value function design and its update in ~\secref{sec:vt_design} and ~\secref{sec:update}, respectively. \figref{fig:alg-framework} summarizes these steps in high level.  

\noindent\textbf{Selection: }
Given a search tree $T$, we denote the molecule nodes as $\Vcal^m(T)$ and reaction nodes as $\Vcal^r(T)$, where the total nodes in $T$ will be $\Vcal(T) = \Vcal^m(T) \cup \Vcal^r(T)$.
The frontier $\Fcal(T) \subseteq \Vcal^m(T)$ contains all the molecule nodes in $T$ that haven't been expanded before.
Since we want to minimize the total cost of the final solution, an ideal option to expand next would be the molecule node which is part of the best synthesis plan.

Suppose we already have a value function oracle $V_t(m|T)$ which tells us that under the current search tree $T$, the cost of the best plan that contains $m$ for synthesizing target $t$. 
 We can use it to select the next node to expand:
\begin{eqnarray}
\textstyle
    m_{next} = \argmin_{m\in\Fcal(T)} V_t(m|T)
\end{eqnarray}
A proper design of such $V_t(m|T)$ would not only improve search efficiency, but can also bring theoretical guarantees. 

\noindent\textbf{Expansion: }
After picking the node $m$ with minimum cost estimation $V_t(m|T)$, we will expand the search tree with $k$ one-step retrosynthesis proposals from $B(m)$. 
Specifically, for each proposed retrosynthesis reaction $(R_i, \Scal_i, c(R_i)) \in B(m)$, we create a reaction node $R=R_i$ under node $m$, and for each molecule $m' \in \Scal_i$, we create a molecule node under the reaction node $R$. This will create an AND-OR stump under node $m$. Unlike in MCTS~\citep{segler2018planning} where multiple calls to $B(\cdot)$ is needed till a terminal state during rollout, here the expansion only requires a single call to the one-step model. 

\noindent\textbf{Update: }
Denote the search tree $T$ after expansion of node $m$ to be $T'$. Such expansion
obtains the corresponding cost information for one-step retrosynthesis. we utilize this more direct information to update $V_t(\cdot|T')$ of all other relevant nodes to provide a more accurate estimation of total cost. 

\subsection{Design of $V_t(m|T)$}
\label{sec:vt_design}

To properly design $V_t(m|T)$, we borrow the idea from A* algorithm.  
A* algorithm is a best-first search algorithm which uses the cost from start $g(\cdot)$ together with the estimation of future cost $h(\cdot)$ to select move. When such estimation is admissible, it will be guaranteed to return the optimal solution.  
Inspired by the A* algorithm, we decompose the value function into two parts:
\begin{equation}
	V_t(m|T) = g_t(m|T) + h_t(m|T)
\label{eq:v=g+t}
\end{equation}
where $g_t(m|T)$ is the cost of current reactions that have happened in $T$, if $m$ should be in the final route, and $h_t(m|T)$ is the estimated cost for future reactions needed to complete such planning. 
Instead of explicitly calculate these two separately, we show an equivalent but simpler way to calculate $V_t(\cdot|T)$ directly. 

Specifically, we first define $V_m(m|\emptyset)$, which is a boundary case of the value function oracle $V$ that simply tells how much cost is needed to synthesize molecule $m$. 
For the simplicity of notation, we denote it as $V_m$. Then we define the \textit{reaction number} function $rn(\cdot|T): \Vcal(T) \mapsto \RR$ that is inspired by proof number but with different purpose: 
\begin{eqnarray}
	rn(R|T) &=& c(R) + \sum_{m \in ch(R)} rn(m|T) \nonumber \\
	rn(m|T) &=& \begin{cases} V_m, \quad\quad\quad\quad\quad\quad\quad m \in \Fcal(T) \\ \min_{R \in ch(m)} rn(R|T), \text{otherwise}\end{cases}
	\label{eq:react_num}
\end{eqnarray}
where $rn(R|T)$ and $rn(m|T)$ calculate for reaction node and molecule node, respectively. 
The reaction number tells the minimum estimated cost needed for a molecule or reaction to happen in the current tree. We further define $pr(u|T): \Vcal(T) \mapsto \Vcal(T)$ to get the parent node of $u$, and $\Acal(u|T)$ be all the ancestors of node $u$. Note that $pr(m|T) \in \Vcal^r(T), \forall m \in \Vcal^m(T)$ and vise versa. Then function $V_t(m|T)$ will be:
\begin{eqnarray}
	V_t(m|T) &=& \sum_{r \in \Acal(m|T)\, \cap\, \Vcal^r(T)} c(r) \nonumber \\
	&+& \sum_{m' \in \Vcal^m(T), pr(m') \in \Acal(m|T) } rn(m'|T) 
	\label{eq:gt}
\end{eqnarray}
The first summation calculates all the reaction cost that has happened along the path from node $m$ to root. Additionally, $\forall R \in \Acal(m|T) \cap \Vcal^{r}(T)$, the child node $m' \in ch(R)$ should also be synthesized, as each such reaction node $R$ is an AND node. This requirement is captured in the second summation of Eq~\eqref{eq:gt}. We can see that 
implicitly $g_t(m|T)$ sums up the cost associated with the reaction nodes in this route related to $m$, and $h_t(m|T)$ takes all the terms related to $V_{\cdot}$ in Eq~\eqref{eq:react_num}.

In~\figref{fig:alg-framework} we demonstrate the calculation of $V_t(m|T)$ with a simple example. 
Notice that we can compute the parts that relevant to $g_t(\cdot|T)$ with existing information. But we can only estimate the part of $h_t(\cdot|T)$ since the required reactions are not in the search tree yet. We will show how to learn this future estimation in~\secref{sec:learning}.
%
%

\subsection{Updating $V_t(m|T)$}
\label{sec:update}

After a node $m$ is expanded, there are several components needed to be updated to maintain the search tree state. 

\noindent\textbf{Update $rn(\cdot|T)$:} Following Eq~\eqref{eq:react_num}, the reaction number for newly created molecule nodes $u$ under the subtree rooted at $m$ will be $V_u$, and the reaction nodes $R \in ch(m)$ will have the cost $c(R)$ added to the sum of reaction numbers in children. After that, all the nodes $u \in \Acal(m|T) \cup \cbr{m}$ would potentially have the reaction number updated following Eq~\eqref{eq:react_num}. Thus this process requires the computation complexity to be $O(\textrm{depth}(T))$. However in our implementation, we can update these nodes in a bottom-up fashion that starts from $m$, and stop anytime when an ancestor node value doesn't change. This would speed up the update.  

\noindent\textbf{Update $V_t(\cdot|T)$:} Let $\Acal'(m|T) \subseteq \rbr{\Acal(m|T) \cup \cbr{m}} \cap \Vcal^m(T)$ be the set of molecule nodes that have reaction number being updated in the stage above. From Eq~\eqref{eq:gt} we can see, for any molecule node $u \in \Fcal(T)$, $V_t(u|T)$ will be recalculated if $\cbr{m': pr(m') \in \Acal(u|T)} \cap \Acal'(m|T) \neq \emptyset$. 

\noindent\textbf{Remark:} The expansion of a node $m$ can potentially affect all other nodes in $\Fcal(T)$ in the worst case. However the expansion of a single molecule node $m$ will only affect another node $v$ in the frontier when it is on the current best synthesis solution that composes $V_t(v|T)$. For the actual implementation, we use efficient caching and lazy propagate mechanism, which will guarantee to only update the $V_t(v|T)$ when it is necessary. The implementation details of both above updates can be found in Appendix~\ref{app:impl_detail}.

\vspace{-1mm}
\subsection{Guarantees on Finding the Optimal Solution}
\vspace{-1mm}

\begin{theorem}
Assuming $V_m$ or its lowerbound is known for all encountered molecules $m$, \algref{alg:main} is guaranteed to return an optimal solution, if the halting condition is changed to ``the total costs of a found route is no larger than $\argmin_{m\in\Fcal(T)} V_t(m)$''.
\label{thm:admissibility}
\end{theorem}
\vspace{-2mm}

The proof can be found in Appendix~\ref{sec:proof}.


\noindent\textbf{Remark 1:} If we define the cost of a reaction to be its negative log-likelihood, then $0$ is the lowerbound of $V_m$ for any molecule $m$. The induced algorithm is guaranteed to find the optimal solution.


\noindent\textbf{Remark 2:} In practice, due to the limited time budget, we prefer the algorithm to return once a solution is found.


\vspace{-1mm}
\subsection{Extension: \modelshort{} on Graph Search Space}
\vspace{-1mm}

We have been mainly illustrating the technique on a tree structured space. 
As the retrosynthesis planning is essentially performend on a directed graph (\ie, certain intermediate molecules may share the same reactants, which may further reduce the actual cost), the above calculation can be extended to the general bipartite graph $G$ with edges connecting $\Vcal^m(G)$ and $\Vcal^r(G)$. Due to the potential existence of loops, the calculation of Eq~\eqref{eq:react_num} will be performed using shortest path algorithm instead. As there will be no negative loops, shortest path algorithm will still converge. By viewing the search space as tree rather than graph, we may possibly find sub-optimal solution due to the repetition in state representation. However, as loopy synthesis is rare in real world,  we mainly focus on the tree structured search in this paper, and will investigate this extension to bipartite graph space search in future work.

\vspace{-2mm}
\section{Estimating $V_m$ from Planning Solutions}
\vspace{-1mm}

\modelshort{} requires the value function oracle $V_m$ to compute $V_t(\cdot|T)$ for expansion node selection. However in practice it is impossible to obtain the exact value of $V_m$ for every molecule $m$. Therefore we try to estimate it from previous planning data.

\subsection{Represention of $V_m$}

To parameterize $V_m$ for any molecule $m$, we first compute its Morgan fingerprint~\citep{rogers2010extended} of radius $2$ with $2048$ bits, and feed it into a single-layer fully connected neural network of hidden dimension $128$, which then outputs a scalar representing $V_m$.

\subsection{Offline Learning of $V_m$}
\label{sec:learning}

Previous work has either used random rollout or human designed heuristics for estimating $V_m$, which may not be accurate enough to guide the search. Instead of learning it online during planning~\citep{silver2017mastering}, we utilize the existing reactions in the training set $\Dcal_{train}$ to train it. 

Specifically, we construct retrosynthesis routes for feasible molecules in $\Dcal_{train}$, where the available set of molecule $\Mcal$ is also given beforehand. The specific construction strategy will be covered in \secref{sec:construct_routes}. The resulting dataset will be $\Rcal_{train} = \cbr{rt_i = (m_i, v_i, R_i, B(m_i))}$, where each tuple $rt_i$ contains the target molecule $m_i$, the best entire route cost $v_i$, the one-step retrosynthesis candidates $B(m_i)$ which also contains the true one-step retrosynthesis $R_i$ used in the planning solution.

The learning of $V_m$ consists of two parts, namely the value fitting which is a regression loss $\Lcal_{reg}(rt_i) = (V_{m_i} - v_i)^2$
and the consistency learning which maintains the partial order relationship between best one-step solution $R_i$ and other solutions $(R_j, \Scal_j, c(R_j)) \in B(m_i)$:
\begin{equation}
	\Lcal_{con}(rt_i, R_j) = \max\cbr{0, v_i + \epsilon - c(R_j) - \sum_{m' \in \Scal_j}V_{m'}}
\end{equation}
where $\epsilon$ is a positive constant margin to ensure $r_i$ has higher priority for expansion than its alternatives even if the value estimates have tolerable noise. 
The overall objective is:
\begin{eqnarray}
	\min_{V_{(\cdot)}} & \EE_{rt_i \sim \Rcal_{train}} \Big[\Lcal_{reg}(rt_i) + \nonumber \\
	 & \lambda \EE_{R_j \sim B(m_i)\setminus \cbr{R_i}} \sbr{\Lcal_{con}(rt_i, R_j)} \Big]
\end{eqnarray}
where $\lambda$ balances these two losses. In experiment we set it to be 1 by default. 

\vspace{-1mm}
\section{Experiments}
\vspace{-1mm}

\begin{table*}[!t]
\centering
\vspace{-1mm}
\begin{tabular}{cccccccc}
\hline
Algorithm      & Retro*  & Retro*-0 & DFPN-E+ & DFPN-E  & MCTS+   & MCTS    & Greedy DFS \\ \hline
Success rate   & 86.84\% & 79.47\%  & 53.68\% & 55.26\% & 35.79\% & 33.68\% & 22.63\%    \\
Time           & 156.58  & 208.58   & 289.42  & 279.67  & 365.21  & 370.51  & 388.15     \\
Shorter routes & 50      & 52       & 59      & 59      & 18      & 14      & 11         \\
Better routes  & 112     & 102      & 22      & 25      & 46      & 41      & 26        \\ \hline
\end{tabular}
\vspace{-1mm}
\caption{Performance summary. Time is measured by the number of one-step model calls, with a hard limit of 500. The number of shorter and better routes are obtained from the comparison against the expert routes, in terms of number of reactions and the total costs.}
\vspace{-2mm}
\label{tbl:summary}
\end{table*}

\subsection{Creating Benchmark Dataset}
\vspace{-1mm}

\subsubsection{USPTO Reaction Dataset}
\vspace{-1mm}

We use the publicly available reaction dataset extracted from United States Patent Office (USPTO) to train one-step model and extract synthesis routes. The whole dataset consists of $\sim 3.8M$ chemical reactions published up to September 2016. For reactions with multiple products, we duplicate them into multiple ones with one product each. After removing the duplications and reactions with wrong atom mappings, we further extract reaction templates with RDChiral~\footnote{\url{https://github.com/connorcoley/rdchiral}} for all reactions and discard those whose reactants cannot be obtained by applying reaction templates to their products. The remaining $\sim 1.3M$ reactions are further split randomly into train/val/test sets following $80\%/10\%/10\%$ proportions.

With reaction data, we train a template-based MLP model~\citep{segler2017neural} for one-step retrosynthesis. Following literature, we formulate the one-step retrosynthesis as a multi-class classification problem, where given a molecule as product, the goal is to predict possible reaction templates. Reactants are obtained by applying the predicted templates to product molecule. There are in total $\sim 380K$ distinct templates. Throughout all experiments, we take the top-$50$ templates predicted by MLP model and apply them on each product to get corresponding reactant lists. 

\subsubsection{Extracting Synthesis Routes}
\label{sec:construct_routes}

To train our value function and quantitatively analyze the predicted routes, we construct synthesis routes based on USPTO reaction dataset and a list of commercially available building blocks from \emph{eMolecules}~\footnote{\url{http://downloads.emolecules.com/free/2019-11-01/}}. ~\emph{eMolecules} consists of $231M$ commercially available molecules that could work as ending points for our searching algorithm. 

Given the list of building blocks, we take each molecule that have appeared in USPTO reaction data and analyze if it can be synthesized by \emph{existing} reactions within USPTO training data. 
For each synthesizable molecule, we choose the shortest-possible synthesis routes with ending points being available building blocks in \emph{eMolecules}. 

We obtain validation and test route datasets with slightly different process. For validation dataset, we first combine train and validation \emph{reaction} dataset, and then repeat aforementioned extraction procedure on the combined dataset. Since we extract routes with more reactions, synthesizable molecules will include those who could not be synthesized with original reactions and those who have shorter routes. We exclude molecules with routes of same length as in training data, and pack the remaining as validation route dataset. We apply similar procedure to test data but make sure that there is no overlap between test and training/validation set. 

We further clean the test route dataset by only keeping the routes whose reactions are all covered by the top-$50$ predictions by the one-step model. To make the test set more challenging, we filter out the easier molecules by running a heuristic-based BFS planning algorithm, and discarding the solved molecules in a fixed time limit. After processing, we obtain $299202$ training routes, $65274$ validation routes, $189$ test routes and the corresponding target molecules.

\subsection{Results}

\begin{figure*}[ht]
\vspace{-2mm}
\centering
\hspace{-6mm}
\includegraphics[width=0.33\textwidth,align=c]{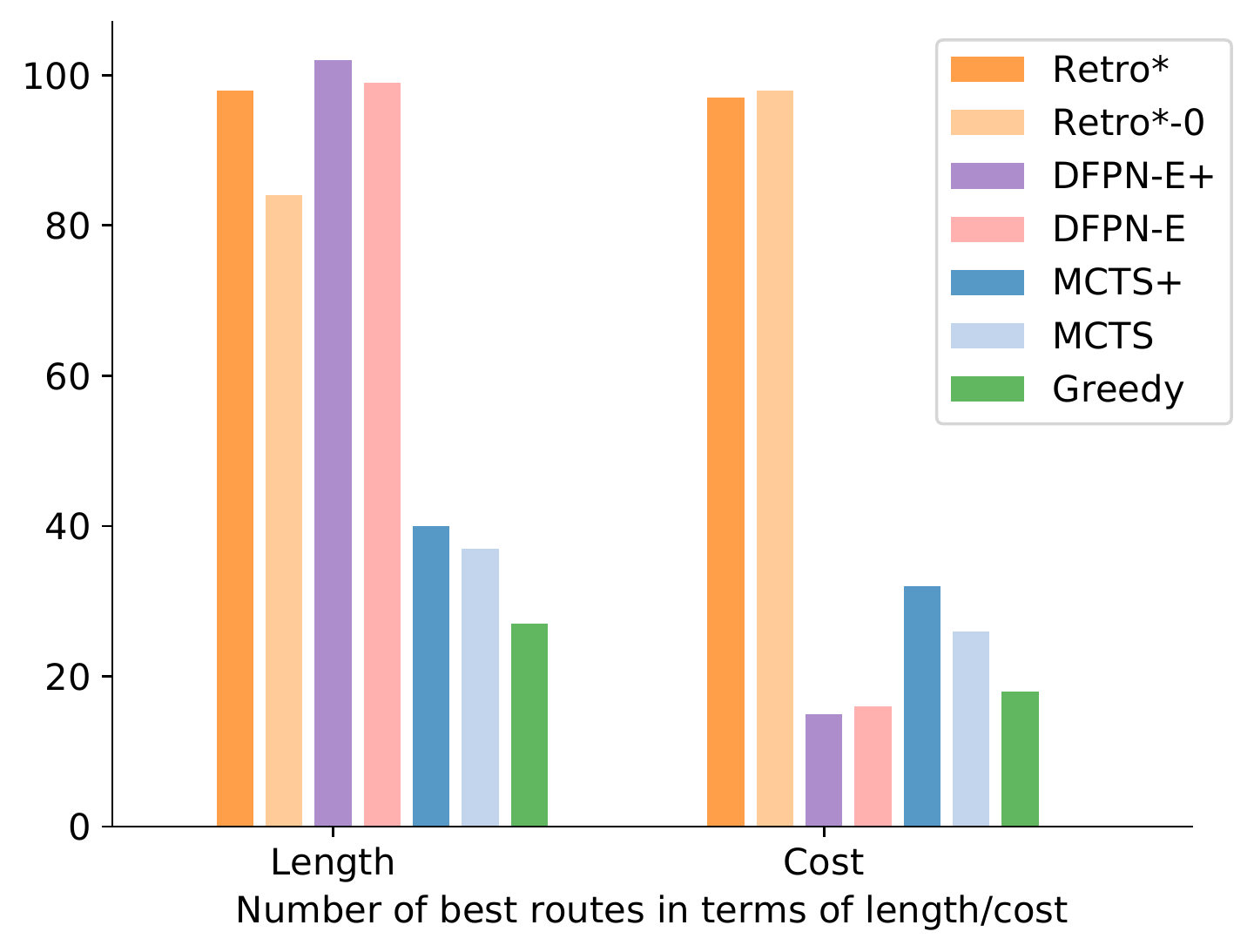}
\hspace{-3mm}
\includegraphics[width=0.68\textwidth,align=c]{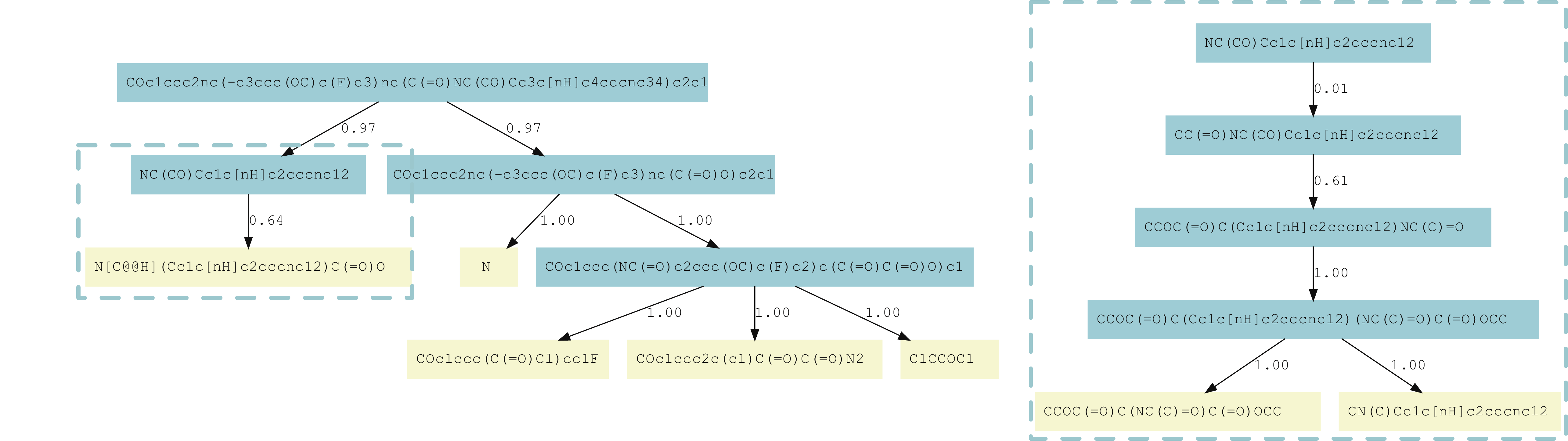}
\vspace{-1mm}
\caption{\textbf{Left}: Counts of the best solutions among all algorithms in terms of length/cost; \textbf{Mid}: Sample solution route from \modelshort{}. Numbers on the edges are the likelihoods of the reactions. Yellow nodes are building blocks; \textbf{Right}: The corresponding dotted box part in the expert route, much longer and less probable than the solution.}
\vspace{-1mm}
\label{fig:route-quality}
\end{figure*}

We compare \modelshort{} against DFPN-E~\citep{kishimoto2019depth}, MCTS~\citep{segler2018planning} and greedy Depth First Search (DFS) on product molecules in test route dataset described in \secref{sec:construct_routes}. Greedy DFS always prioritizes the reaction with the highest likelihood. MCTS is implemented with PUCT, where we used the reaction probability provided by the one-step model as the prior to bias the search.

We measure both route quality and planning efficiency to evaluate the algorithm.
To measure the quality of a solution route, we compare its total cost as well as its length, \ie~number of reactions in the route. The cost function is defined as the negative log-likelihood of the reaction. Therefore, minimizing the total costs is equivalent to maximizing the likelihood of the route. To measure planning effiency, we use the number of calls to the one-step model ($\approx 0.3s$ per call) as a surrogate of time (since it will occupy $>99\%$ of running time) and compare the success rate under the same time limit. 

\noindent\textbf{Performance summary:}
The performances of all algorithms are summarized in \tabref{tbl:summary}. Under the time limit of $500$ one-step calls, \modelshort{} solves $31\%$ more test molecules than the second best method, DFPN-E. Among all the solutions given by \modelshort{}, $50$ of them are shorter than expert routes, and $112$ of them are better in terms of the total costs.
We also conduct an ablation study to understand the importance of the learning component in \modelshort{} by evaluating its non-learning version \modelshort{}-0. \modelshort{}-0 is obtained from \modelshort{} by setting $V_m$ to $0$, which is a lowerbound of any valid values. Comparing to baseline methods, \modelshort{}-0 is also showing promising results. However, it is outperformed by \modelshort{} by $6\%$ in terms of success rate, demonstrating the performance gain brought by learning from previous planning experience. 

To find out whether MCTS and DFPN-E can benefit from the learned value function oracle $V_m$ in \modelshort{}, we replace the reward estimation by rollout in MCTS and the proof number initialization in DFPN-E by the same $V_m$, calling the strengthened algorithms MCTS+ and DFPN-E+. Value function helps MCTS as expected due to having a value estimate with less variance than rollout. The performance of DFPN-E is not improved because we don’t have a good initialization of the disproof number.

\begin{figure}[ht]
\vspace{-2mm}
\centering
\includegraphics[width=0.45\textwidth]{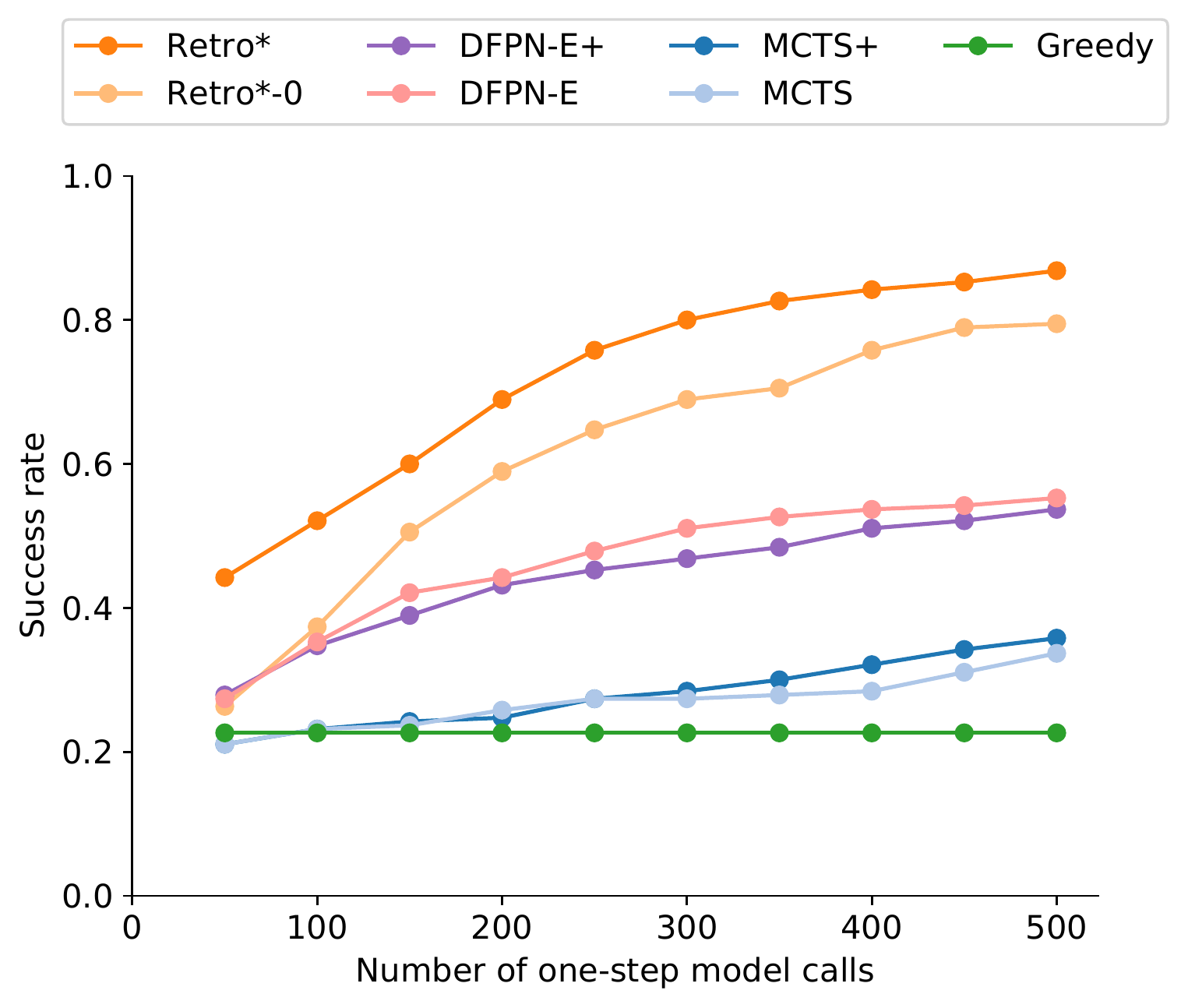}
\vspace{-2mm}
\caption{Influence of time limit on performance.}
\label{fig:succ-rate}
\end{figure}

\noindent\textbf{Influence of time limit:}
To show the influence of time limit on performance, we plot the success rate against the number of one-step model calls in \figref{fig:succ-rate}. We can see that \modelshort{} not only outperforms baseline algorithms by a large margin at the beginning, but also is improving faster than the baselines, enlarging the performance gap as the time limit increases.

\noindent\textbf{Solution quality:} 
To evaluate the overall solution quality, for each test molecule, we collect solutions from all algorithms, and compare the route lengths and costs (see \figref{fig:route-quality}-left). We only keep the best routes (could be multiple) for each test molecule, and count the number of best routes in total for each method. We find that in terms of total costs, \modelshort{} produces $4\times$ more best routes than the second best method. Even for the length metric, which is not the objective \modelshort{} is optmizing for, it still achieves about the same performance as the best method.

As a demonstration for \modelshort{}'s ability to find high-quality routes, we illustrate a sample solution in \figref{fig:route-quality}-mid, where each node represents a molecule. The target molecule corresponds to the root node, and the building blocks are in yellow. The numbers on the edges indicates the likelihoods of successfully producing the corresponding reactions in realworld. The expert route provided shares the exactly the same first reaction and the same right branch with the route found by our algorithm. However, the left branch (\figref{fig:route-quality}-right) is much longer and less probable than the corresponding part of the solution route, as shown in the dotted box region in \figref{fig:route-quality}-mid. Please refer to Appendix~\ref{sec:sample-solutions} for more sample solution routes and search tree visualizations.

\section{Conclusion}

In this work, we propose \modelshort{}, a learning-based retrosynthetic planning algorithm for efficiently finding high-quality routes. \modelshort{} is able to utilize previous planning experience to bias the search on unseen molecules towards promising directions. We also propose a systematic approach for creating a retrosynthesis dataset from publicly available reaction datasets and novel metrics for evaluating solution routes without involving human experts. Experiments on realworld benchmark dataset demonstrate our algorithm's significant improvement over existing methods on both planning efficiency and solution quality.

\section*{Acknowledgements}

We thank Junhong Liu, Wei Yang and Yong Liu for helpful discussions. This work is supported in part by NSF grants CDS\&E-1900017 D3SC, CCF-1836936 FMitF, IIS-1841351, CAREER IIS-1350983, CNS-1704701, ONR MURI grant to L.S.

\bibliography{bibfile}
\bibliographystyle{icml2020}

\clearpage

\onecolumn

\appendix

\section{Implementation details}
\label{app:impl_detail}

\begin{figure}[ht]
\centering
\includegraphics[width=\textwidth]{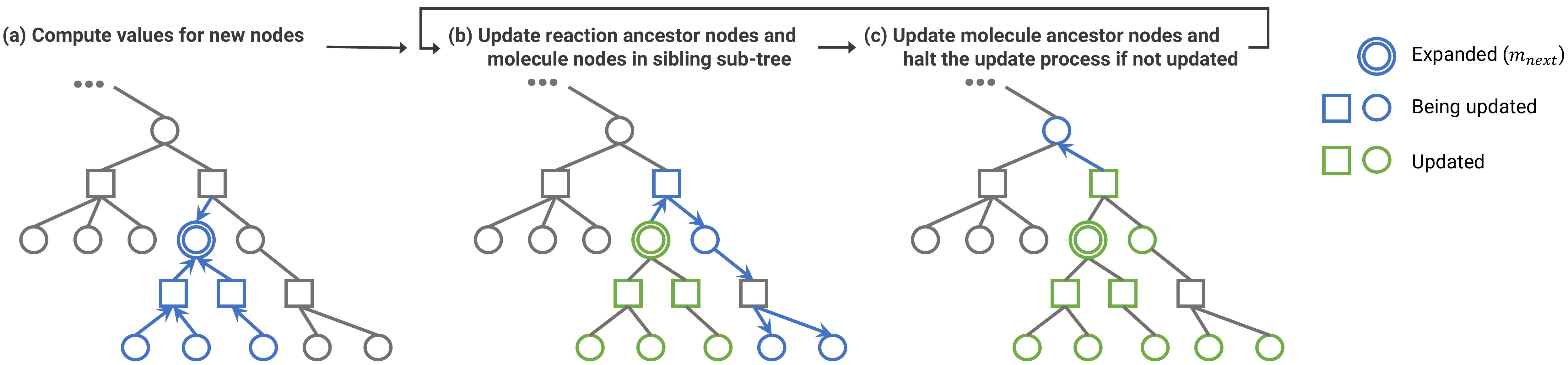}
\caption{Illustration for the update process. Three phases correspond to line~\ref{ln:start}-\ref{ln:new-values}, line~\ref{ln:r-update-start}-\ref{ln:r-update-end}, and line~\ref{ln:m-update-start}-\ref{ln:end} in \algref{alg:update}.}
\label{fig:update}
\end{figure}

In this section we describe the algorithm details in the update phase of \modelshort{}. The goal of the update phase is to compute the up-to-date $V_t(m|T)$ for every molecule node $m\in\Fcal(T)$. To implement efficient update, we need to cache $V_t(m|T)$ for all $m\in\Vcal^m(T)$. Note that from Eq~\eqref{eq:gt}, we can observe the fact that sibling molecule nodes have the same $V_t(m|T)$, \ie~$V_t(m_a|T)=V_t(m_b|T)$ if $pr(m_a|T) = pr(m_b|T)$. Therefore instead of storing the value of $V_t(m|T)$ in every molecule node $m$, we store the value in their common parent via defining $V_t(R|T)=V_t(m|T)$ if $R=pr(m|T)$ for every reaction node $R\in\Vcal^r(T)$.

In our implementation, we cache $V_t(R|T)$ for all reaction nodes $R\in\Vcal^r(T)$ and cache $rn(v|T)$ for all nodes $v\in\Vcal(T)$. Caching values in this way would allow us to visit each related node only once for minimal update.

\begin{algorithm2e}[ht]
\For{$i\gets1$ \KwTo $k$}{\label{ln:start}
    \For{$m \in \Scal_i$}{
        $rn(m) \gets V_m$\;
    }
    $rn(R_i) \gets c(R_i) + \sum_{m \in \Scal_i} rn(m)$\;
    $V_t(R_i) \gets V_t(pr(m_{next})) - rn(m_{next}) + rn(R_i)$\;
}
$new\_rn \gets \min_{i\in\{1,2,\cdots,k\}} rn(R_i)$\;
$delta \gets new\_rn - rn(m_{next})$\;
$rn(m_{next}) \gets new\_rn$\;\label{ln:new-values}
$m_{current} \gets m_{next}$\;\label{ln:bottom-up-start}
\While{$delta \neq 0$ and $m_{current}$ is not root}{\label{ln:stop-criteria}
    $R_{current} \gets pr(m_{current})$\;\label{ln:r-update-start}
    $rn(R_{current}) \gets rn(R_{current}) + delta$\;
    $V_t(R_{current}) \gets V_t(R_{current}) + delta$\;
    \For{$m \in ch(R_{current})$}{
        \If{$m$ is not $m_{current}$}{
            $\mathtt{UpdateSibling}(m, delta)$\;\label{ln:sib}
        }
    }\label{ln:r-update-end}
    $m_{current} \gets pr(R_{current})$\;\label{ln:m-update-start}
    $delta = 0$\;
    \If{$rn(R_{current}) < rn(m_{current})$}{
        $delta \gets rn(R_{current}) - rn(m_{current})$\;
        $rn(m_{current}) \gets rn(R_{current})$\;\label{ln:end}
    }
}
\caption{$\mathtt{Update}(m_{next}, \{R_i, \Scal_i, c(R_i)\}_{i=1}^k)$ \protect\footnotemark}
\label{alg:update}
\end{algorithm2e}

\footnotetext{For clarity, we omit the condition on $T$ in the notations.}

The update function is summarized in \algref{alg:update} and illustrated in \figref{fig:update}, which takes in the expanded node $m_{next}$ and the expansion result $\{R_i, \Scal_i, c(R_i)\}_{i=1}^k$, and performs updates to affected nodes. We first compute the values for new reactions according to Eq~\eqref{eq:react_num} and \eqref{eq:gt} in line~\ref{ln:start}-\ref{ln:new-values}. Then we update the ancestor nodes of $m_{next}$ in a bottom-up fashion in line~\ref{ln:bottom-up-start}-\ref{ln:end}. We also update the molecule nodes in the sibling sub-trees in line~\ref{ln:sib} and \algref{alg:update-sibling}.

\begin{algorithm2e}[ht]
$rn(m|T) \gets rn(m|T) + delta$\;
\For{$R \in ch(m|T)$}{
    \For{$m' \in ch(R|T)$}{
        $\mathtt{UpdateSibling}(m', delta)$\;
    }
}
\caption{$\mathtt{UpdateSibling}(m, delta)$}
\label{alg:update-sibling}
\end{algorithm2e}

Our implementation visits a node only when necessary. When updating along the ancestor path, it immediately stops when the influence of the expansion vanishes (line~\ref{ln:stop-criteria}). When updating a single node, we use a $O(1)$ delta update by leveraging the relations derived from Eq~\eqref{eq:react_num} and \eqref{eq:gt}, avoiding a direct computation which may require $O(k)$ or $O(\textrm{depth}(T))$ summations.

\section{Guarantees on finding the optimal solution}
\label{sec:proof}

Since \modelshort{} is a variant of the A* algorithm, we can leverage existing results to prove the theoretical guarantees for \modelshort{}. In this section, we first state the assumptions we make, and then prove the admissibility~(\thmref{thm:admissibility}) of \modelshort{}.

The theoretical results in this paper build upon the assumption that we can access $\hat{V}_m$, which is a lowerbound for $V_m$ for all molecules $m$. Note that this is a weak assumption, since we know $0$ is a universal lowerbound for $V_m$.

As we describe in Eq~\eqref{eq:v=g+t}, $V_t(m|T)$ can be decomposed into $g_t(m|T)$ and $h_t(m|T)$, where $g_t(m|T)$ is the exact cost of the partial route through $m$ which is already in the tree, and $h_t(m|T)$ is the future costs for frontier nodes in the route which is a summation of a series of $V_m$s. In practice we use $\hat{V}_m$ in the summation, and arrive at $\hat{h}_t(m|T)$, which is a lowerbound of $h_t(m|T)$,~\ie~the following lemma.

\begin{lemma}
Assuming $V_m$ or its lowerbound is known for all encountered molecules $m$, then the approximated future costs $\hat{h}_t(m|T)$ in \modelshort{} is a lowerbound of true $h_t(m|T)$.
\label{lm:admissibility}
\end{lemma}

We re-state the admissibility result~(\thmref{thm:admissibility}) in the main text and prove it with existing results in A* literature.

\begingroup
\def\thetheorem{\ref{thm:admissibility}}
\begin{theorem}
\textbf{(Admissibility)} Assuming $V_m$ or its lowerbound is known for all encountered molecules $m$, \algref{alg:main} is guaranteed to return an optimal solution, if the halting condition is changed to ``the total costs of a found route is no larger than $\argmin_{m\in\Fcal(T)} V_t(m)$''.
\end{theorem}
\addtocounter{theorem}{-1}
\endgroup

\begin{proof}
Combine \lemref{lm:admissibility} and Theorem 1 in the original A* paper \citep{hart1968formal}.
\end{proof}

\section{Sample search trees and solution routes}
\label{sec:sample-solutions}

In this section, we present two examples of the solution routes and the corresponding search trees for target molecule $A$ and $B$ produced by \modelshort{}. 

\begin{figure}[ht]
    \centering
    \includegraphics[width=0.9\textwidth]{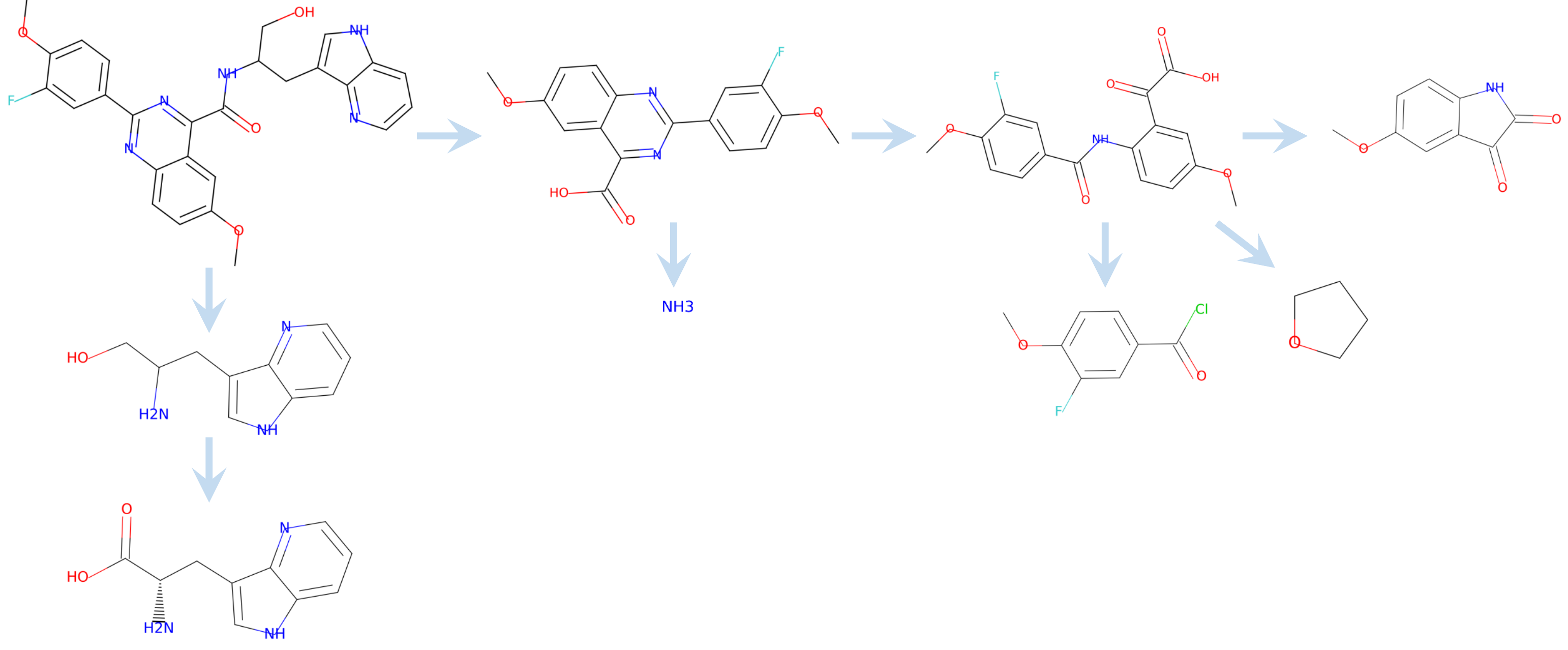}
    \includegraphics[width=0.85\textwidth]{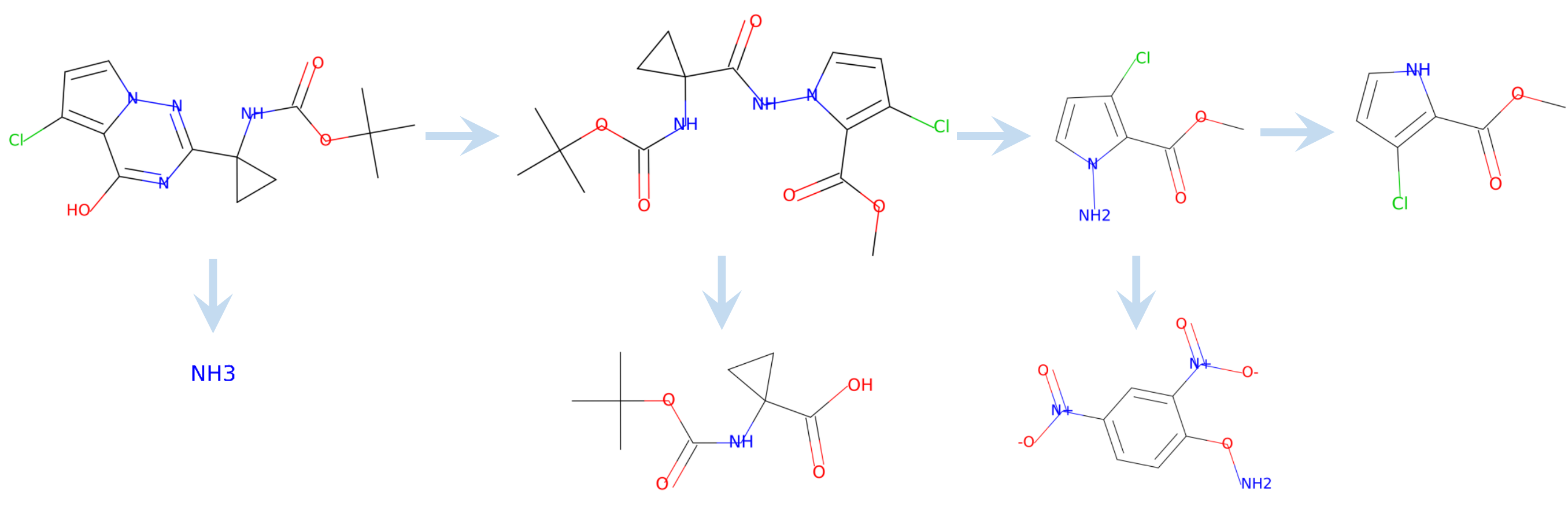}
    \caption{Top/bottom: solution route produced by \modelshort{} for molecule $A/B$. Edges point from the same product molecule to the reactant molecules represent an one-step chemical reaction.}
    \label{fig:route_ex12}
\end{figure}

Solution route for target molecule $A/B$ is illustrated in the top/bottom sub-figure of \figref{fig:route_ex12}, where a set of edges pointing from the same product molecule to reactant molecules represents an one-step chemical reaction. Molecules on the leaf nodes are all available.

\begin{figure}[ht]
    \centering
    \includegraphics[width=\textwidth]{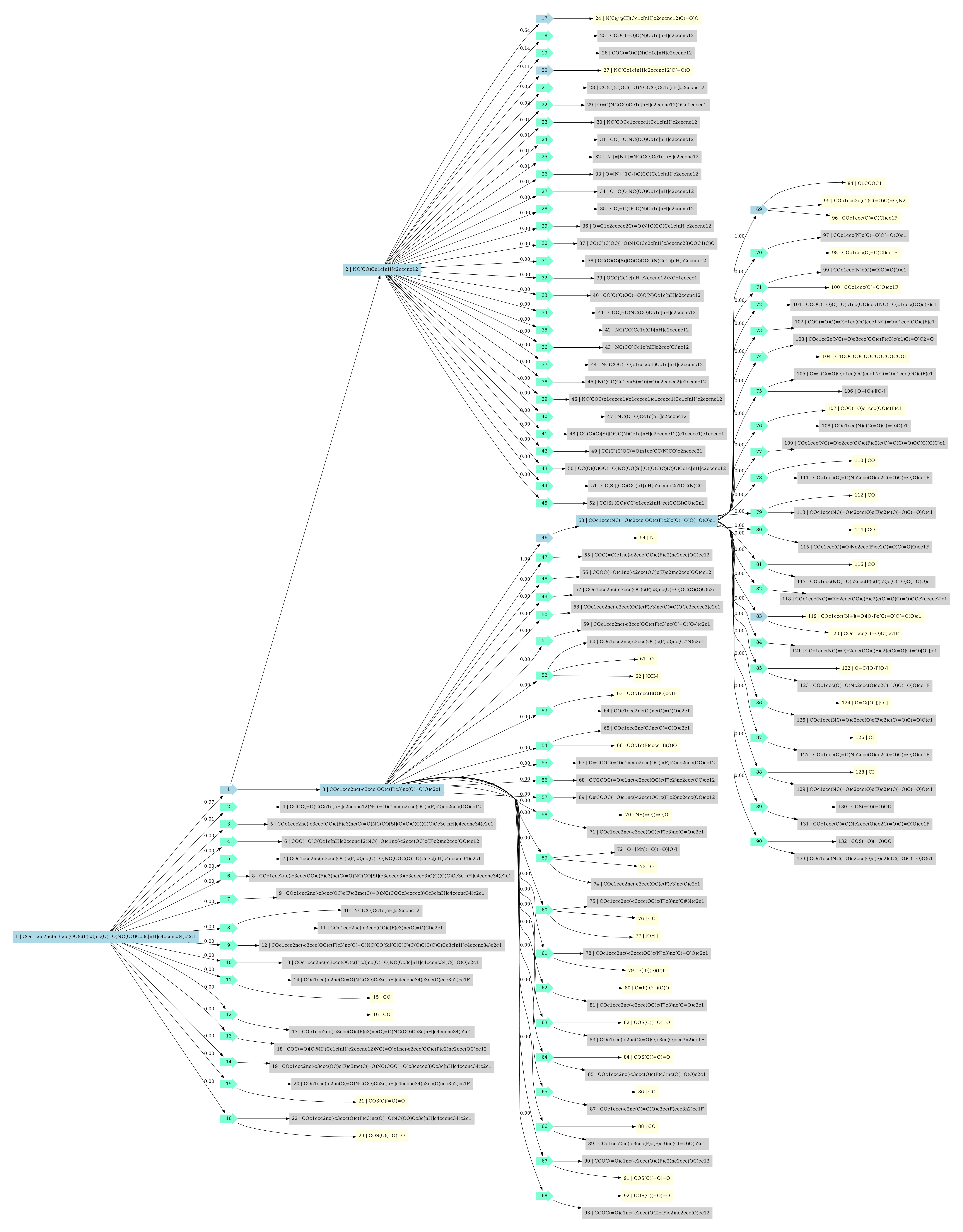}
    \caption{Search tree produced by \modelshort{} for molecule $A$. Reactangular boxes/arrows represent molecules/reactions. Yellow/grey/blue indicate available/unexpanded/solved molecules. Numbers on the edges are the probabilities produced by the one-step model.}
    \label{fig:search_tree_ex1}
\end{figure}

\begin{figure}[ht]
    \centering
    \includegraphics[width=0.6\textwidth]{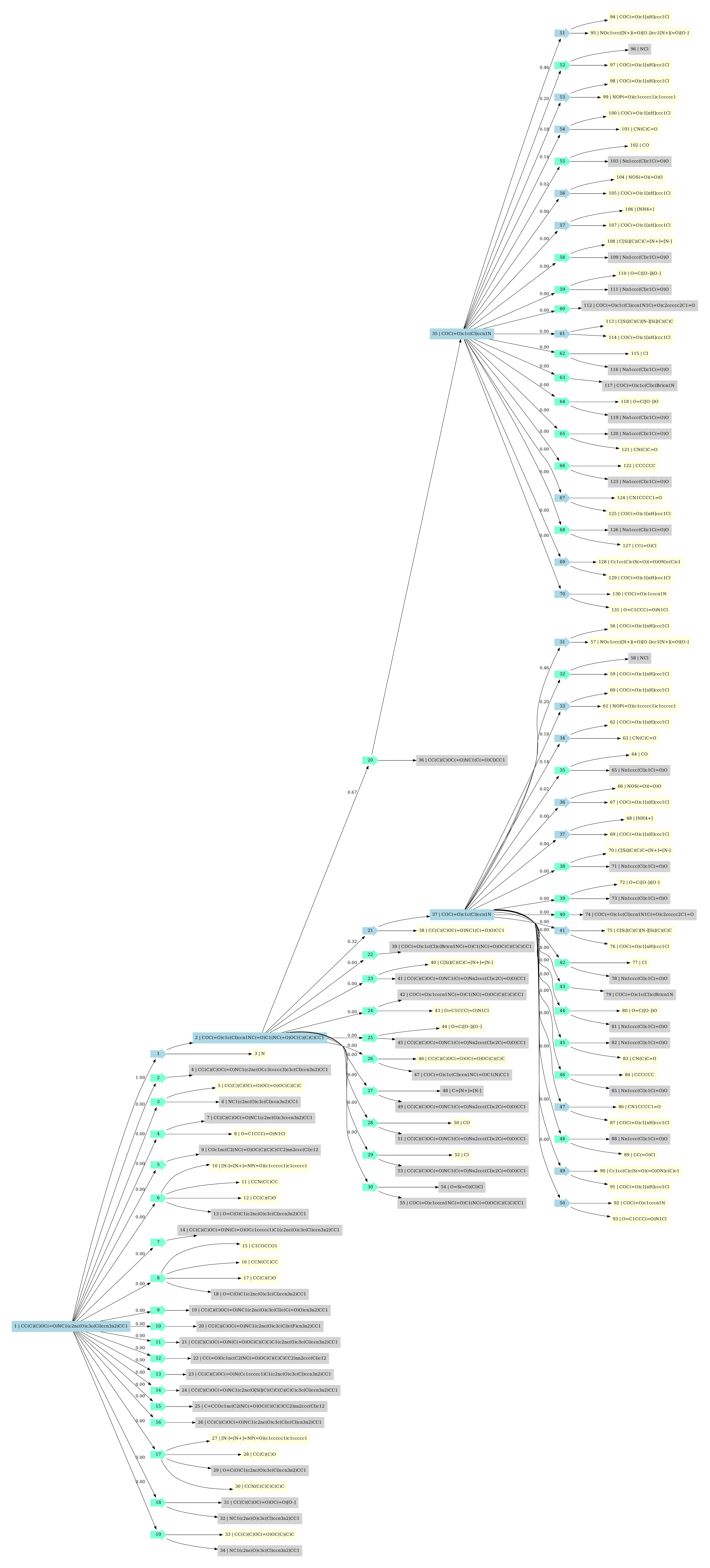}
    \caption{Search tree produced by \modelshort{} for molecule $B$. Reactangular boxes/arrows represent molecules/reactions. Yellow/grey/blue indicate available/unexpanded/solved molecules. Numbers on the edges are the probabilities produced by the one-step model.}
    \label{fig:search_tree_ex2}
\end{figure}

The search trees for molecule $A$ and $B$ are illustrated in \figref{fig:search_tree_ex1} and \figref{fig:search_tree_ex2}. We use reactangular boxes to represent molecules. Yellow/grey/blue boxes indicate available/unexpanded/solved molecules. Reactangular arrows are used to represent reactions. The numbers on the edges pointing from a molecule to a reaction are the probabilities produced by the one-step model. Due to space limit, we only present the minimal tree which leads to a solution.

\section{\modelshort{} for hierarchical task planning}
\label{sec:htn}

As a general planning algorithm, Retro* can be applied to other machine learning problems as well, including theorem proving~\citep{yang2019learning} and hierarchical task planning~\citep{erol1996hierarchical} (or HTP), etc. Below, we conduct a synthetic experiment on HTP to demonstrate the idea. In the experiment, we are trying to search for a plan to complete a target task. The tasks (OR nodes) can be completed with different methods, and each method (AND nodes) requires a sequence of subtasks to be completed. Furthermore, each method is associated with a nonnegative cost. The goal is to find a plan with minimum total cost to realize the target task by decomposing it recursively until all the leaf task nodes represent primitive tasks that we know how to execute directly.
As an example, to travel from home in city $A$ to hotel in city $B$, we can take either \texttt{flight}, \texttt{train} or \texttt{ship}, each with its own cost. For each method, we have subtasks such as home $\rightarrow$ airport $A$, \texttt{flight}($A\rightarrow B$), and airport $B$ $\rightarrow$ hotel. These subtasks can be further realized by several methods.

As usual, we want to find a plan with small cost in limited time which is measured by the number of expansions of task nodes. We use the optimal halting condition as stated in theorem~\ref{thm:admissibility}. We compare our algorithms against DFPN-E, the best performing baseline. The results are summarized in \tabref{tbl:htn-succ} and~\ref{tbl:htm-ar}.

\begin{table}[h]
\centering
\begin{tabular}{|l|l|l|l|l|l|}
\hline
\multicolumn{1}{|c|}{{Time Limit}} &
  \multicolumn{1}{c|}{{15}} &
  \multicolumn{1}{c|}{{20}} &
  \multicolumn{1}{c|}{{25}} &
  \multicolumn{1}{c|}{{30}} &
  \multicolumn{1}{c|}{{35}} \\ \hline
Retro*   & .67 & .91 & .96 & .98 & 1.  \\ \hline
Retro*-0 & .50 & .86 & .95 & .98 & .99 \\ \hline
DFPN-E   & .02 & .33 & .74 & .93 & .97 \\ \hline
\end{tabular}
\caption{Success rate (higher is better) vs time limit.}
\label{tbl:htn-succ}
\end{table}

\begin{table}[h]
\centering
\begin{tabular}{l|l|l|l|l|}
\cline{2-5}
 & \multicolumn{1}{c|}{{Alg}} & \multicolumn{1}{c|}{{Retro*}} & \multicolumn{1}{c|}{{Retro*-0}} & \multicolumn{1}{c|}{{DFPN-E}} \\ \cline{2-5} 
 & Avg. AR                            & 1                                    & 1                                      & 1.5                                  \\ \cline{2-5} 
 & Max. AR                            & 1                                    & 1                                      & 3.9                                  \\ \cline{2-5} 
\end{tabular}
\caption{AR = Approximation ratio (lower is better), time limit=35.}
\label{tbl:htm-ar}
\end{table}

As we can see, in terms of success rate, \modelshort{} is slightly better than \modelshort{}-0, and both of them are significantly better than DFPN-E. In terms of solution quality, we compute the approximation ratio (= solution cost / ground truth best solution cost) for every solution, and verify the theoretical guarantee in theorem~\ref{thm:admissibility} on finding the best solution.

\section{Related Works}
\label{sec:related}

Reinforcement learning algorithms (without planning) have also been considered for the retrosynthesis problem. \citet{schreck2019learning} leverages self-play experience to fit a value function and uses policy iteration for learning an expansion policy. It is possible to combine it with a planning algorithm to achieve better performance in practice.

Learning to search from previous planning experiences has been well studied and applied to Go~\citep{silver2016mastering, silver2017mastering}, Sokoban~\citep{guez2018learning} and path planning~\citep{chen2020learning}. Existing methods cannot be directly applied to the retrosynthesis problem since the search space is more complicated, and the traditional representation where a node corresponds to a state is highly inefficient, as we mentioned in the discussion on MCTS in previous sections.

\end{document}